\documentclass[10pt,twocolumn,letterpaper]{article}

\usepackage{cvpr}      
\usepackage{amsmath}
\usepackage[dvipsnames]{xcolor}

\definecolor{cvprblue}{rgb}{0.21,0.49,0.74}
\usepackage[pagebackref,breaklinks,colorlinks,citecolor=cvprblue]{hyperref}

\usepackage{listings}
\usepackage[accsupp]{axessibility}

\def\paperID{4206} 
\def\confName{CVPR}
\def\confYear{2024}

\newcommand*{\affaddr}[1]{#1} 
\newcommand*{\affmark}[1][*]{\textsuperscript{#1}}

\usepackage[symbol]{footmisc}

\title{UltrAvatar: A Realistic Animatable 3D Avatar Diffusion Model with Authenticity Guided Textures}

\def\authorBlock{
    Mingyuan Zhou\footnote{}~~\affmark[1], \
    Rakib Hyder\footref{note1}~~\affmark[1], \
    Ziwei Xuan~~\affmark[1], \
    Guojun Qi \affmark[1,]\affmark[2] \\
    \small{\affaddr{\affmark[1]OPPO US Research Center, InnoPeak Technology, Inc., USA, ~~}}\small{\affaddr{\affmark[2]Westlake University, China}}\\
    {\tt\small \{mingyuan.zhou,rakib.hyder,ziwei.xuan\}@innopeaktech.com}, {\tt\small\ {guojunq}@gmail.com}
}
\author{\authorBlock}

\begin{document}

\twocolumn[{%
\renewcommand\twocolumn[1][]{#1}%
\maketitle
\begin{center}
    \centering
    \captionsetup{type=figure}
    \includegraphics[width=1.0\textwidth,height=6cm]{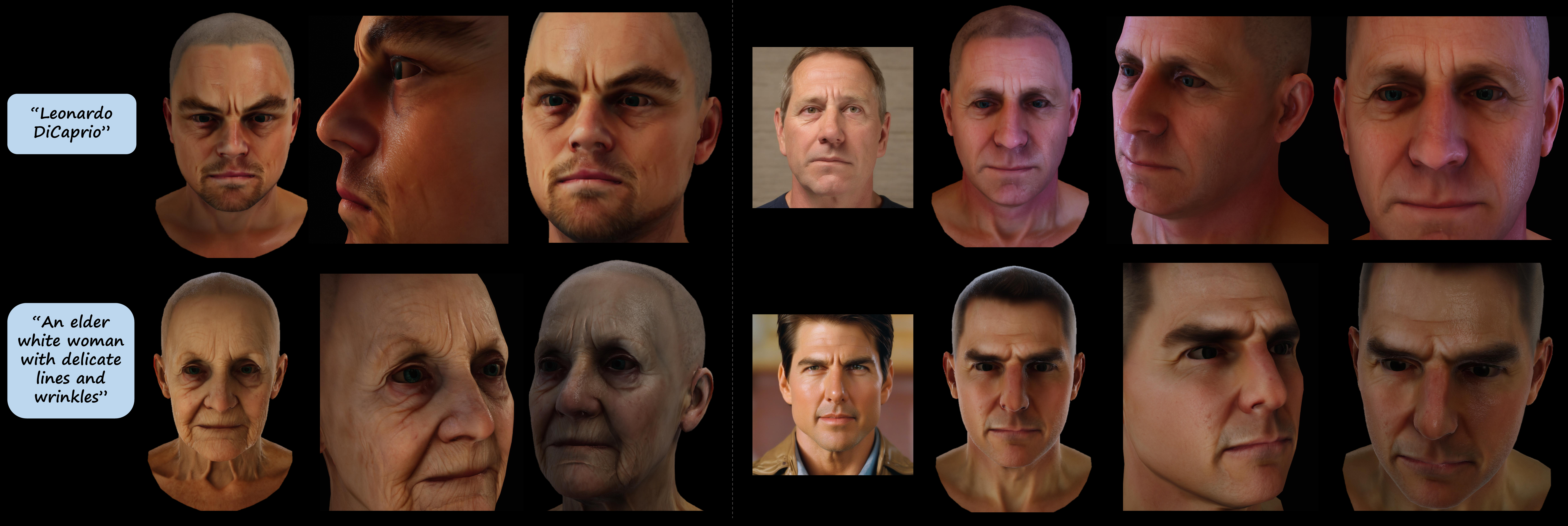}
    \captionof{figure}{\textbf {UltrAvatar}. Our method takes a text prompt or a single image as input to generate realistic animatable 3D Avatars with PBR textures, which are compatible with various rendering engines, our generation results in a wide diversity, high quality, and excellent fidelity. }
    \label{fig:teaser}
    
\end{center}%
}]

\footnotetext[1]{\label{note1} Equal contribution.}

\begin{abstract}

Recent advances in 3D avatar generation have gained significant attention. These breakthroughs aim to produce more realistic animatable avatars, narrowing the gap between virtual and real-world experiences. Most of existing works employ Score Distillation Sampling (SDS) loss, combined with a differentiable renderer and text condition, to guide a diffusion model in generating 3D avatars. However, SDS often generates over-smoothed results with few facial details, thereby lacking the diversity compared with ancestral sampling. On the other hand, other works generate 3D avatar from a single image, where the challenges of unwanted lighting effects, perspective views, and inferior image quality make them difficult to reliably reconstruct the 3D face meshes with the aligned complete textures. In this paper, we propose a novel 3D avatar generation approach termed UltrAvatar with enhanced fidelity of geometry, and superior quality of physically based rendering (PBR) textures without unwanted lighting. To this end, the proposed approach presents a diffuse color extraction model and an authenticity guided texture diffusion model. The former removes the unwanted lighting effects to reveal true diffuse colors, so that the generated avatars can be rendered under various lighting conditions. The latter follows two gradient-based guidances for generating PBR textures to render diverse face-identity features and details better aligning with 3D mesh geometry. We demonstrate the effectiveness and robustness of the proposed method, outperforming the state-of-the-art methods by a large margin in the experiments.


\end{abstract}

\section{Introduction}
\label{sec:intro}

Generating 3D facial avatars is of significant interest in the communities of both computer vision and computer graphics. Recent advancements in deep learning have greatly enhanced the realism of AI-generated avatars. Although multi-view 3D reconstruction methods, such as Multi-View Stereo (MVS) \cite{seitz2006comparison} and Structure from Motion (SfM) \cite{schonberger2016structure}, have facilitated avatar generation from multiple images captured at various angles, generating realistic 3D avatars from few images, like a single view taken by a user or particularly generated from text prompts, is significantly challenging due to the limited visibility, unwanted lighting effects and inferior image quality.

Previous works have attempted to overcome these challenges by leveraging available information contained in the single view image. For example, \cite{zheng2022avatar, li2017learning, flametex} focused on estimating parameters of the 3D Morphable Model (3DMM) model by minimizing landmark loss and photometric loss, and other approaches train a self-supervised network to predict 3DMM parameters
\cite{sanyal2019learning,feng2021learning,danvevcek2022emoca,zielonka2022towards}. These methods are sensitive to occlusions and lighting conditions, leading to susceptible 3DMM parameters or generation of poor quality textures. Moreover, many existing works \cite{danvevcek2022emoca,zielonka2022towards,zheng2022avatar,feng2021learning} rely on prefixed texture basis \cite{gerig2018morphable} to generate facial textures. Although these textures are often reconstructed jointly with lighting parameters, the true face colors and skin details are missing in the underlying texture basis and thus are unable to be recovered. Alternatively, other works \cite{an2023panohead,yin2023nerfinvertor,hong2022headnerf,athar2022rignerf} employ neural radiance rendering field (NeRF) to generate 3D Avatar, but they are computationally demanding and not amenable to mesh-based animation of 3D avatars. They also may lack photo-realism when being rendered from previously unobserved perspectives. 

Generative models \cite{gecer2019ganfit,lattas2020avatarme,gecer2020synthesizing,rowan2023text2face,zhang2023text, papantoniou2023relightify} designed for 3D avatars have shown promising to generate consistent 3D meshes and textures. However, these models do not account for unwanted lighting effects that prevent access to true face colors and could result in deteriorated diffuse textures. On the other hand, some works use the SDS loss \cite{poole2022dreamfusion,zhang2023dreamface,zhang2023text} to train a 3D avatar by aligning the rendered view with the textures generated by a diffusion model. The SDS may lead to the oversmoothed results that lack the diversity in skin and facial details compared with the original 2D images sampled from the underlying diffusion model.

To overcome these challenges, we propose a novel approach to create 3D animatable avatars with better diffuse colors and more detailed skin textures. First, our approach can take either a text prompt or a single face image as input. The text prompt is fed into a generic diffusion model to create a face image, or a single face image can also be input into our framework. It is well known that separating lighting from the captured colors in a single image is intrinsically challenging. To obtain high quality textures that are not contaminated by the unwanted lighting, our key observation is that the self-attention block in the diffusion model indeed captures the lighting effects, enabling us to reveal the true diffuse colors by proposing a diffuse color extraction (DCE) model to robustly eliminate the lighting from the texture of the input image.

In addition, we propose to train an authenticity guided texture diffusion model (AGT-DM) that is able to generate high-quality complete facial textures that align with the 3D face meshes. Two gradient guidances are presented to enhance the resultant 3D avatars -- a photometric guidance and an edge guidance that are added to classifier-free diffusion sampling process. This can improve the diversity of the generated 3D avatars with more subtle high-frequency details in their facial textures across observed and unobserved views.


The key contributions of our work are summarized below.

\begin{itemize}[leftmargin=*]
\item We reveal the relationship between the self-attention features and the lighting effects, enabling us to propose a novel model for extracting diffuse colors by
removing lighting effects in a single image. Our experiments demonstrate this is a robust and effective approach, suitable for tasks aimed at removing specular spotlights and shadows. 

\item We introduce an authenticity guided diffusion model to generate PBR textures. It can provide high-quality complete textures that well align with 3D meshes without susceptible lighting effects. The sampling process follows two gradient-based guidances to retain facial details unique to each identity, which contributes to the improved generation diversity.

\item We build a novel 3D avatar generative framework, UltrAvatar, upon the proposed DCE model and the AGT-DM. Our experiments demonstrate high quality diverse 3D avatars with true colors and sharp details (see Fig.~\ref{fig:teaser}). 




\end{itemize}	 

\begin{figure*}[t]
  \centering
   \includegraphics[width=0.96\linewidth]{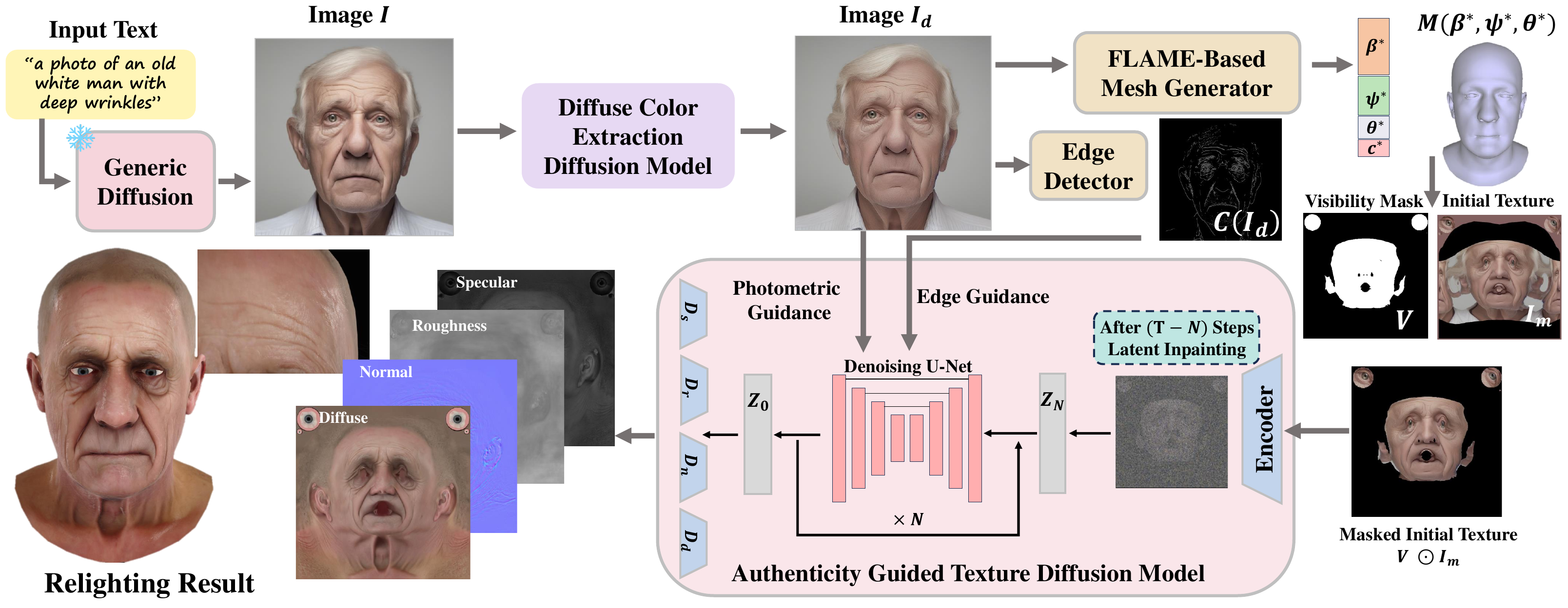}

   \caption{\textbf{The Overview of UltrAvatar.} First, we feed a text prompt into a generic diffusion model (SDXL~\cite{podell2023sdxl}) to produce a face image. Alternatively, the face image can also be used directly as input into our framework. Second, our DCE model takes the face image to extract its diffuse colors $I_d$ by eliminating lighting. The $I_d$ is then passed to the mesh generator and the edge detector to generate the 3D mesh, camera parameters and the edge image. With these predicted parameters, the initial texture and the corresponding visibility mask can be created by texture mapping. Lastly, we input the masked initial texture into our AGT-DM to generate the PBR textures. A relighting result using the generated mesh and PBR textures is shown here. }
   \label{fig:pipeline}

\end{figure*}

\section{Related Work}
\label{sec:related_work}

\paragraph{Image-to-Avatar Generation:}

Initially, avatar generation was predominantly reliant on complex and costly scanning setups, limiting its scalability~\cite{guo2019relightables,lombardi2018deep,alexander2010digital,borshukov2005realistic}. This challenge has shifted towards utilizing common image inputs like a single photo~\cite{feng2021learning,danvevcek2022emoca,zielonka2022towards} or video~\cite{cao2022authentic,gao2022reconstructing,grassal2022neural,zheng2022avatar}. Additionally, the representation of 3D heads has diversified, ranging from the mesh-based parametric models~\cite{zheng2022avatar, li2017learning} to the fluid neural implicit functions like NeRFs~\cite{an2023panohead, yin2023nerfinvertor, hong2022headnerf}. The introduction of advanced neural networks, especially Generative Adversarial Networks (GANs)~\cite{karras2020analyzing} has led to the embedding of 3D attributes directly into these generative models \cite{or2022stylesdf,chan2022efficient,deng2022gram} and the use of generative model as a prior to generate 3D avatars~\cite{lattas2023fitme, gecer2019ganfit, lattas2020avatarme, gecer2020synthesizing} etc. Some techniques~\cite{athar2022rignerf, ding2023diffusionrig, papantoniou2023relightify} can also create the riggable avatars from a single image. Nevertheless, these existing methods rely on the provided images for mesh and texture generation, encountering issues with misalignment between texture and mesh due to errors in mesh generation and discrepancies between visible and invisible regions, and challenges in achieving highly precise diffuse textures due to small specular spots and shadows. Our proposed method adeptly addresses and mitigates these challenges, ensuring more consistent and accurate results.

\paragraph{Text-to-3D Generation:}

Text-to-3D generation is a popular research topic that builds on the success of text-to-image models~\cite{ramesh2022hierarchical,ramesh2021zero,rombach2022high,nichol2022glide}. DreamFusion~\cite{poole2022dreamfusion},  Magic3D~\cite{lin2023magic3d}, Latent-NeRF~\cite{metzer2023latent}, AvatarCLIP~\cite{hong2022avatarclip}, ClipFace~\cite{aneja2023clipface}, Rodin~\cite{wang2023rodin}, DreamFace~\cite{zhang2023dreamface} etc., use the text prompt to guide the 3D generation. Most of these approaches use SDS loss to maintain consistency between the images generated by the diffusion model and the 3D object. However, SDS loss significantly limits the diversity of generation. Our approach upholds the powerful image generation capabilities from diffusion models trained on large-scale data, facilitating diversity. Simultaneously, it ensures a high degree of fidelity between the textual prompts and the resulting avatars without depending on SDS loss.

\paragraph{Guided Diffusion Model:}

A salient feature of diffusion models lies in their adaptability post-training, achieved by guiding the sampling process to tailor outputs. The concept of guided diffusion has been extensively explored in a range of applications, encompassing tasks like image super-resolution~\cite{gao2023implicit,chung2022diffusion,saharia2022image}, colorization~\cite{saharia2022palette,chung2022improving}, deblurring~\cite{chung2022diffusion,whang2022deblurring}, and style-transfer~\cite{gal2022image,kumari2023multi,kwon2022diffusion}. Recent studies have revealed that the diffusion U-Net's intermediate features are rich in information about the structure and content of generated images~\cite{tumanyan2023plug,hertz2022prompt,kwon2022diffusion,preechakul2022diffusion,avrahami2023spatext}. We discover that the attention features can represent lighting in the image and propose a method to extract the diffuse colors from a given image. Additionally, we incorporated two guidances to ensure the authenticity and realism of the generated avatars.
\section{The Method}

An overview of our framework is illustrated in Fig.~\ref{fig:pipeline}. We take a face image as input or use the text prompt to generate a view $I$ of the avatar with a diffusion model. 
Then, we introduce a DCE model to recover diffuse colors by eliminating unwanted lighting from the generated image. This process is key to generating high quality textures without being deteriorated by lighting effects such as specularity and shadows. This also ensures the generated avatars can be properly rendered under new lighting conditions. We apply a 3D face model (e.g., a 3DMM model) to generate the mesh aligned with the resultant diffuse face image. Finally, we apply AGT-DM with several trained decoders to generate PBR textures, including diffuse colors, normal, specular, and roughness textures. This complete set of PBR textures can align well with the 3D mesh, as well as preserve the face details unique to individual identity.

\subsection{Preliminaries}
\label{sec:background}

Diffusion models learn to adeptly transform random noise with condition $y$ into a clear image by progressively removing the noise. These models are based on two essential processes. The forward process initiates with a clear image $x_0$ and incrementally introduces noise, culminating in a noisy image $x_T$, and the backward process works to gradually remove the noise from $x_T$, restoring the clear image $x_0$. The stable diffusion (SD) model \cite{rombach2022high, podell2023sdxl} operates within the latent space $z = E(x)$ by encoding the image $x$ to a latent representation. The final denoised RGB image is obtained by decoding the latent image through $x_0 = D(z_0)$. To carry out the sequential denoising, the network $\epsilon_\theta$ is rigorously trained to predict noise at each time step $t$ by following the objective function:

\begin{align}
   \min_{\theta}\;  \mathbb{E}_{z \sim E(x), t, \epsilon \sim \mathcal{N}(0, 1)}||\epsilon - \epsilon_{\theta}(z_t, t, \tau (y))||_2^2.
\end{align}

where the $\tau(\cdot)$ is the conditioning encoder for an input condition $y$, such as a text embedding, $z_t$ represents the noisy latent sample at the time step $t$. 
The noise prediction model in SD is based on the U-Net architecture, where each layer consists of a residual block, a self-attention block, and a cross-attention block, as depicted in Fig. \ref{fig:rm_model}. At a denoising step $t$, the features $\phi_{t}^{l-1}$ from the previous $(l-1)$-th layer are relayed to the residual block to produce the res-features $f_t^l$. Within the self-attention block, the combined features $(\phi_{t}^{l-1} + f_t^l)$ through the residual connection are projected to the query features $q_t^l$, key features $k_t^l$ and value features $v_t^l$. The above res-features $f_t^l$ contributes to the content of the generated image and the attention features hold substantial information that contributes to the overall structure layout, which are normally used in image editing \cite{tumanyan2023plug,hertz2022prompt,kwon2022diffusion,preechakul2022diffusion}. 

Diffusion models possess the pivotal feature of employing guidance to influence the reverse process for generating conditional samples. Typically, classifier guidance can be applied to the score-based models by utilizing a distinct classifier. Ho \emph{et al.} \cite{ho2021classifier} introduce the classifier-free guidance technique, blending both conditioned noise prediction $\epsilon_{\theta}(z_t, t, \tau (y))$  and unconditioned noise prediction $\epsilon_{\theta}(z_t, t,\varnothing)$, to extrapolate one from another, 
\begin{align}
    \tilde{\epsilon}_\theta(z_t,t,\tau (y)) = \omega \epsilon_\theta(z_t,t,\tau (y)) + (1-\omega) \epsilon_\theta(z_t,t,\varnothing).
\end{align}
where $\varnothing$ is the embedding of a null text and $\omega$ is the guidance scale.



\begin{figure}[t]
  \centering
   \includegraphics[width=0.95\linewidth]{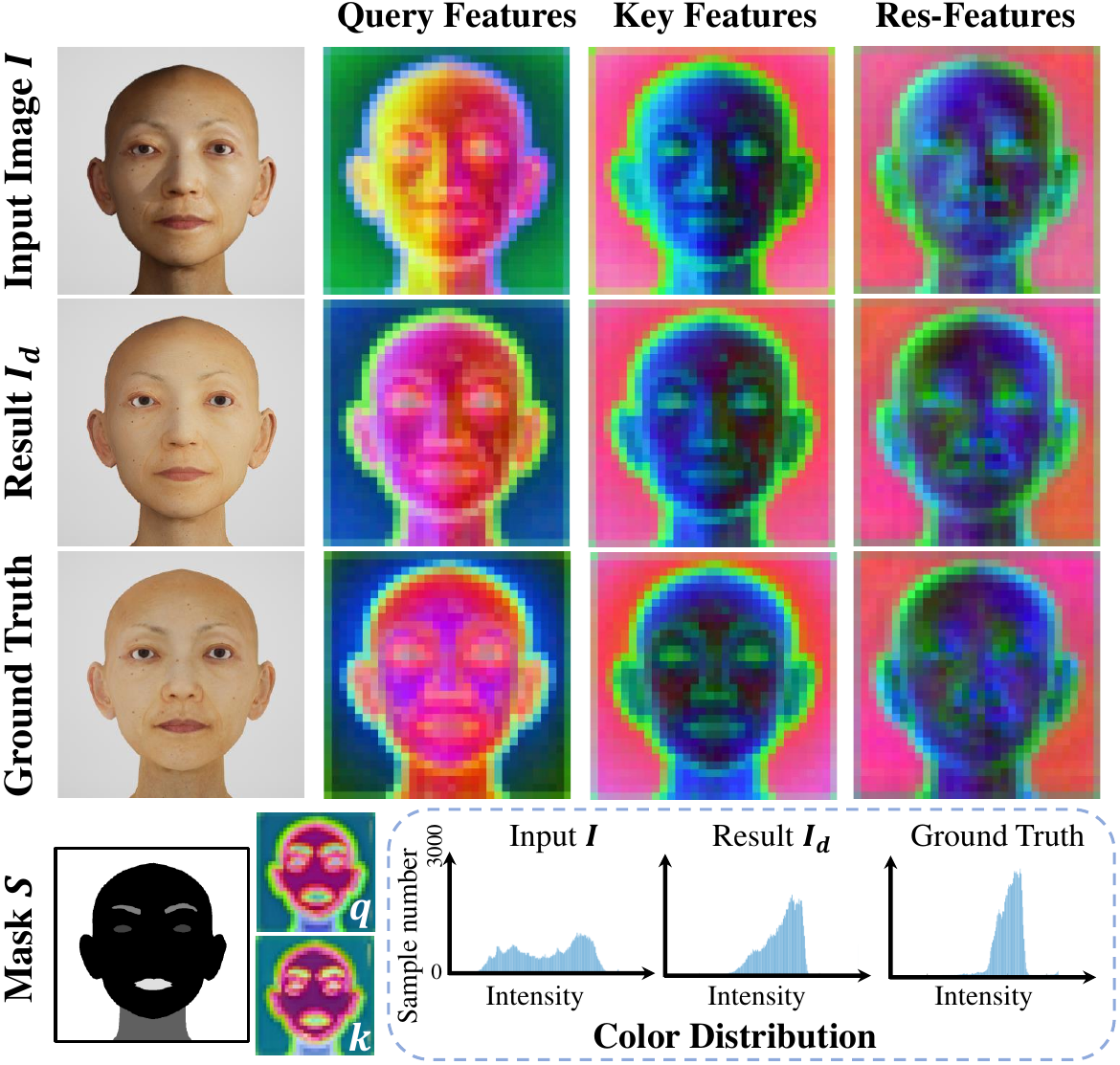}

   \caption{ \textbf{Features Visualization.} We render a high-quality data with PBR textures under a complex lighting condition to image $I$, and also render its corresponding ground truth diffuse color image. We input the $I$ to our DCE model to produce result $I_d$. The $S$ is the semantic mask. We apply DDIM inversion and sampling on these images and extract the features. To visualize the features, we apply PCA on the extracted features to check the first three principal components. The attention features and res-features shown here are all from the $8$-th layer at upsampling layers in the U-Net at time step $101$. From the extracted query and key features of $I$, we can clearly visualize the lighting. The colors and extracted query and key features of the result $I_d$ closely match those from the ground truth image, which demonstrates our method effectively removes the lighting. All res-features do not present too much lighting. We also show the color distributions of these three images, illustrating that the result $I_d$ can eliminate shadows and specular points, making its distribution similar to the ground truth. }
   \label{fig:atten}
   \vspace{-3.00mm} 
\end{figure}

\subsection{Diffuse Color Extraction via Diffusion Features}

Our approach creates a 3D avatar from a face image $I$ that is either provided by a user or generated by a diffusion model  \cite{rombach2022high,podell2023sdxl} from a text prompt. To unify the treatment of those two cases, the DDIM inversion \cite{song2020denoising, dhariwal2021diffusion} with non-textual condition is applied to the image that results in a latent noise $z_T^I$ at time step $T$ from which the original image $I$ is then reconstructed through the backward process. This gives rise to a set of features from the diffusion model.

The given image $I$, no matter user-provided or SD-generated, typically contains shadows, specular highlights, and lighting effects that are hard to eliminate. To render a relightable and animatable 3D avatar, it usually requires a diffuse texture map with these lighting effects removed from the image, which is a challenging task. For this, we make a key observation that reveals the relation between the self-attention features and the lighting effects in the image, and introduce a DCE model to eliminate the lighting.


First, we note that the features $f_t^l$ in each layer contain the RGB details as discussed in \cite{si2023freeu, tumanyan2023plug}. The self-attention features $q_t^l$ and $k_t^l$ reflect the image layout, with similar regions exhibiting similar values.
Beyond this, our finding is that the variations in these self-attention features $q_t^l$ and $k_t^l$ indeed reflect the variations caused by the lighting effects such as shading, shadows, and specular highlights 
within a semantic region. This is illustrated in Fig.~\ref{fig:atten}. This concept is readily graspable. Consider a pixel on the face image, its query features ought to align with the key features from the same facial part so that its color can be retrieved from the relevant pixels. With the lighting added to the diffuse image, the query features must vary in the same way as the variation caused by the lighting effects. In this way, the lighted colors could be correctly retrieved corresponding to the lighting pattern -- shadows contribute to the colors of nearby shadowed pixels, while highlights contribute to the colors of nearby highlighted ones.

\begin{figure}[t]
  \centering
   \includegraphics[width=1.0\linewidth]{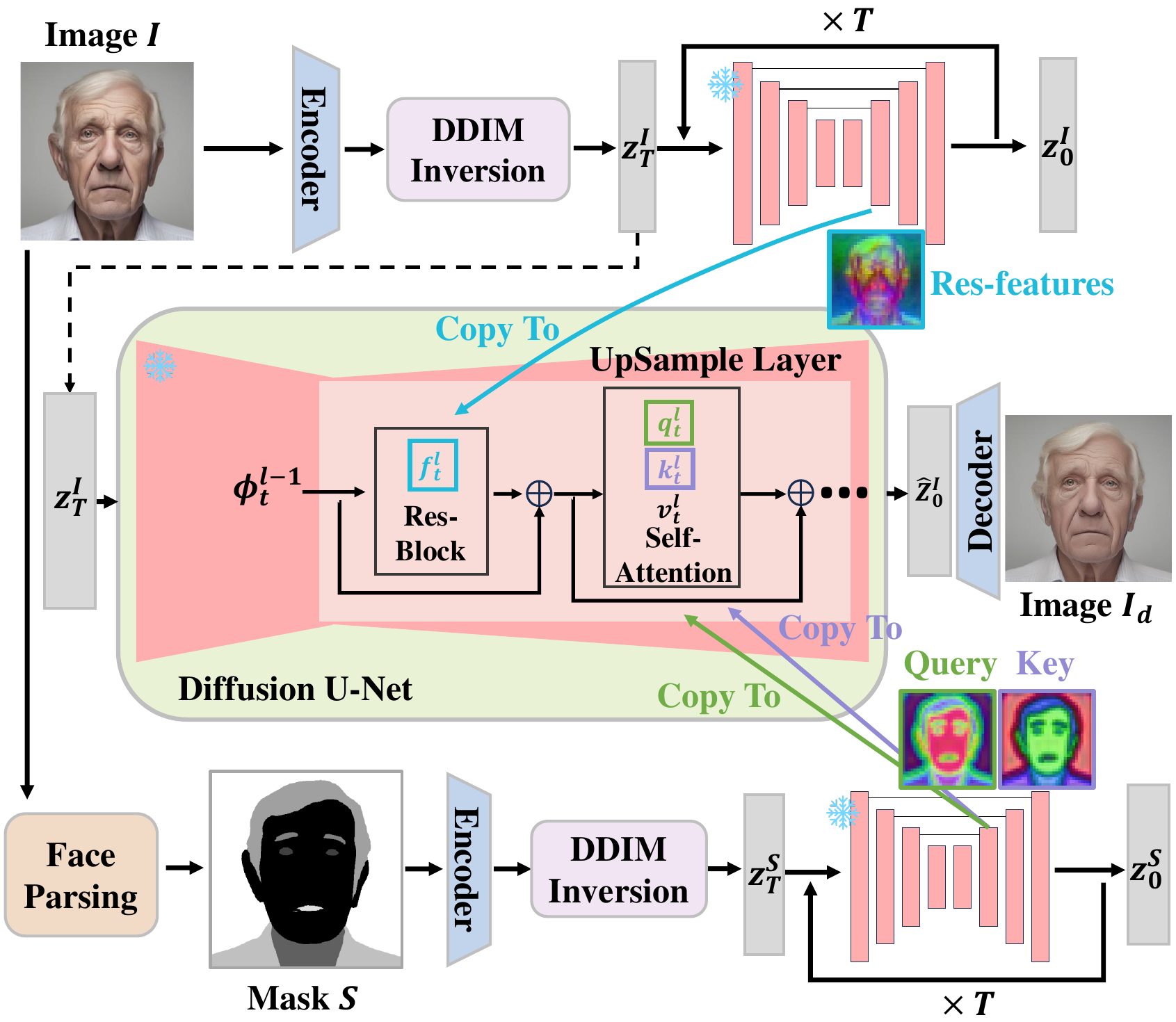}
   \caption{ \textbf{DCE Model.} The input image $I$ is fed to the face parsing model to create the semantic mask $S$. We apply DDIM inversion on the $I$ and $S$ to get initial noise $z^I_T$ and $z^S_T$, then we progressively denoise the $z^I_T$ and $z^S_T$ to extract and preserve the res-features and attention features separately. Lastly, we progressively denoise the $z^I_T$ one more time, copying the res-features and attention features from storage at certain layers (as discussed in Sec. \ref{sec:Experiment}) during sampling to produce ${\hat z}_0^I$, the final result $I_d$ will be generated from decoding the ${\hat z}_0^I$. }
   \label{fig:rm_model}
   \vspace{-3.00mm} 
\end{figure}

To eliminate lighting effects, one just needs to remove the variation in the self-attention (query and key) features while still keeping these features aligned with the semantic structure. Fig.~\ref{fig:rm_model} summarizes the idea. Specifically, first we choose a face parsing model to generate a semantic mask $S$ for the image $I$. The semantic mask meets the above two requirements since it perfectly aligns with the semantic structure by design and has no variation within a semantic region. Then we apply the DDIM inversion to $S$ resulting in a latent noise $z^S_T$ at time step $T$, and obtain the self-attention features of $S$ via the backward process starting from $z^S_T$ for further replacing $q_t^l$ and $k_t^l$ of the original $I$. Since the semantic mask has uniform values within a semantic region, the resultant self-attention features are hypothesized to contain no lighting effects (see Fig.~\ref{fig:atten}), while the face details are still kept in the  features $f_t^l$ of the original image $I$. Thus, by replacing the query and key features $q_t^l$ and $k_t^l$ with those from the semantic mask in the self-attention block, we are able to eliminate the lighting effects from $I$ and keep its diffuse colors through the backward process starting from the latent noise $z_T^I$ used for generating $I$. 

This approach can be applied to eliminate lighting effects from more generic images other than face images, and we show more results in the supplementary material.


\subsection{3D Avatar Mesh Generation}

We employ the FLAME~\cite{li2017learning} model as our geometry representation of 3D avatars. FLAME is a 3D head template model, which is trained from over $33,000$ scans. It is characterized by the parameters for identity shape $\beta$ , facial expression $\psi$ and pose parameters $\theta$. With these parameters. FLAME generates a mesh $M(\beta,\psi,\theta)$ consisting 5023 vertices and 9976 faces, including head, neck, and eyeballs meshes. 
We adopt MICA~\cite{zielonka2022towards} for estimating shape code $\beta^{*}$ of the FLAME model from the diffuse image $I_d$, which excels in accurately estimating the neutral face shape and is robust to expression, illumination, and camera changes. We additionally apply EMOCA~\cite{danvevcek2022emoca} to obtain the expression code $\psi^{*}$, pose parameters $\theta^{*}$ and camera parameters $c^{*}$, which are employed for subsequent 3D animation/driving applications. Note that we do not use the color texture generated by the EMOCA combining the FLAME texture basis. It cannot accurately present the true face color, lacks skin details and contains no PBR details, such as diffuse colors, normal maps, specularity, and roughness textures, which can be derived below with our AGT-DM.

\subsection{Authenticity Guided Texture Diffusion Model}

Given the current estimated mesh $M(\beta^*,\psi^*,\theta^*)$, camera parameters $c^{*}$ and the lighting-free face image $I_d$, one can do the texture mapping of the latter onto the mesh, and then project the obtained mesh texture to an initial texture UV map $I_m$. Since $I_d$ is only a single view of the face, the resultant $I_m$ is an incomplete UV texture map of diffuse colors, and we use $V$ to denote its visible mask in the UV coordinates. The UV texture map may also not perfectly align with the mesh due to the imperfect estimation of face pose, expression and camera parameters by EMOCA.

To address the above challenges, we train an AGT-DM that can 1) inpaint the partially observed texture UV map $I_m$ by $T-N$ steps to fill in the unobserved regions, 2) improve the alignment between the texture map and the 3D mesh by leveraging the texture diffusion model as a prior, and 3) preserve the identity and facial details by employing two guidance signals based on the photometric and edge details. Moreover, the model can output more PBR details beyond the diffuse color textures, including normal, specular and roughness maps from the given $I_m$ and $V$. 


To this end, we use the online 3DScan dataset~\cite{3DStore} that consists of high-quality 3D face scans alongside multiple types of PBR texture maps including diffuse colors, normal maps, specularity and roughness textures. We process this dataset (details in the supplementary material) as a training dataset to train a texture diffusion model where the U-net of the original SD is finetuned over the ground truth diffuse UV maps from the dataset. To generate PBR textures, the SD encoder is frozen and the SD decoder is finetuned for each type of PBR textures (specularity, roughness and normal), except for the the PBR decoder for diffuse texture, $D_d$ which directly inherits from the original SD decoder. Then we can use the fine-tuned texture diffusion model to inpaint the masked diffuse color map $V \odot I_m $ for the first $T-N$ steps to get $Z_N$. We denoise $Z_N$ for the rest $N$ steps to achieve output $Z_0$. Because the training dataset has ideally aligned meshes and texture details, the resultant texture diffusion model can improve the alignment between the output PBR textures and meshes by denoising the noisy texture latent $Z_N$ generated from latent inpainting.

To further enhance the PBR textures with more facial details, we employ two guidance terms to guide the sampling process of the texture diffusion model. The first is the photometric guidance $G_P$ with the following energy function,
{\small
\begin{align}
    & G_P = \omega_{photo} ||V_d \odot (R(M(\beta^*,\psi^*,\theta^*),D_d(z_t),c^*) - I_d) ||_2^2 \notag\\
    & + \omega_{lpips} L_{lpips} (V_d \odot (R(M(\beta^*,\psi^*,\theta^*),D_d(z_t),c^*))  ,V_d \odot I_d )
\end{align} }%
where $V_d$ is the mask over the visible part of rendered face, as shown in Fig. \ref{fig:AGT}, and the $R(\cdot)$ is a differential renderer of the avatar face based on the current estimate of the mesh $M$, the diffuse color texture map $D_d(z_t)$ at a diffusion time step $t$, the $L_{lpips}(.)$ is the perceptual loss function (LPIPS \cite{zhang2018unreasonable}). The minimization of this photometric energy will align the rendered image with the original image.

The second is the edge guidance with the following edge energy function,
{\small
\begin{align}
   G_E = ||V_d \odot ( C(R(M(\beta^*,\psi^*,\theta^*),D(z_t),c^*)) - C(I_d) )||_2^2
\end{align}}%
where $C(\cdot)$ is the canny edge detection function \cite{canny1986computational}. While the edges contain high-frequency details, as shown in Fig. \ref{fig:pipeline}, this guidance will help retain the facial details such as wrinkles, freckles, pores, moles and scars in the image $I_d$, making the generated avatars look more realistic with high fidelity. 






We integrate the two guidances through the gradients of their energy functions into the sampling of classifier-free guidance below,   
\begin{align}
    \tilde{\epsilon}_\theta(z_t,t,\tau (y)) &= \omega \epsilon_\theta(z_t,t,\tau (y)) + (1-\omega) \epsilon_\theta(z_t,t,\varnothing) \notag\\ 
    & \qquad  + \omega_p \nabla_{z_t} G_P + \omega_e \nabla_{z_t} G_E .
\label{eq:gf}
\end{align}
We demonstrate the effectiveness in Fig~\ref{fig:AGT}. 


\section{Experiments}
\label{sec:Experiment}
\subsection{Setup and Baselines}

\begin{figure*}[t]
  \centering
   \includegraphics[width=0.98\linewidth]{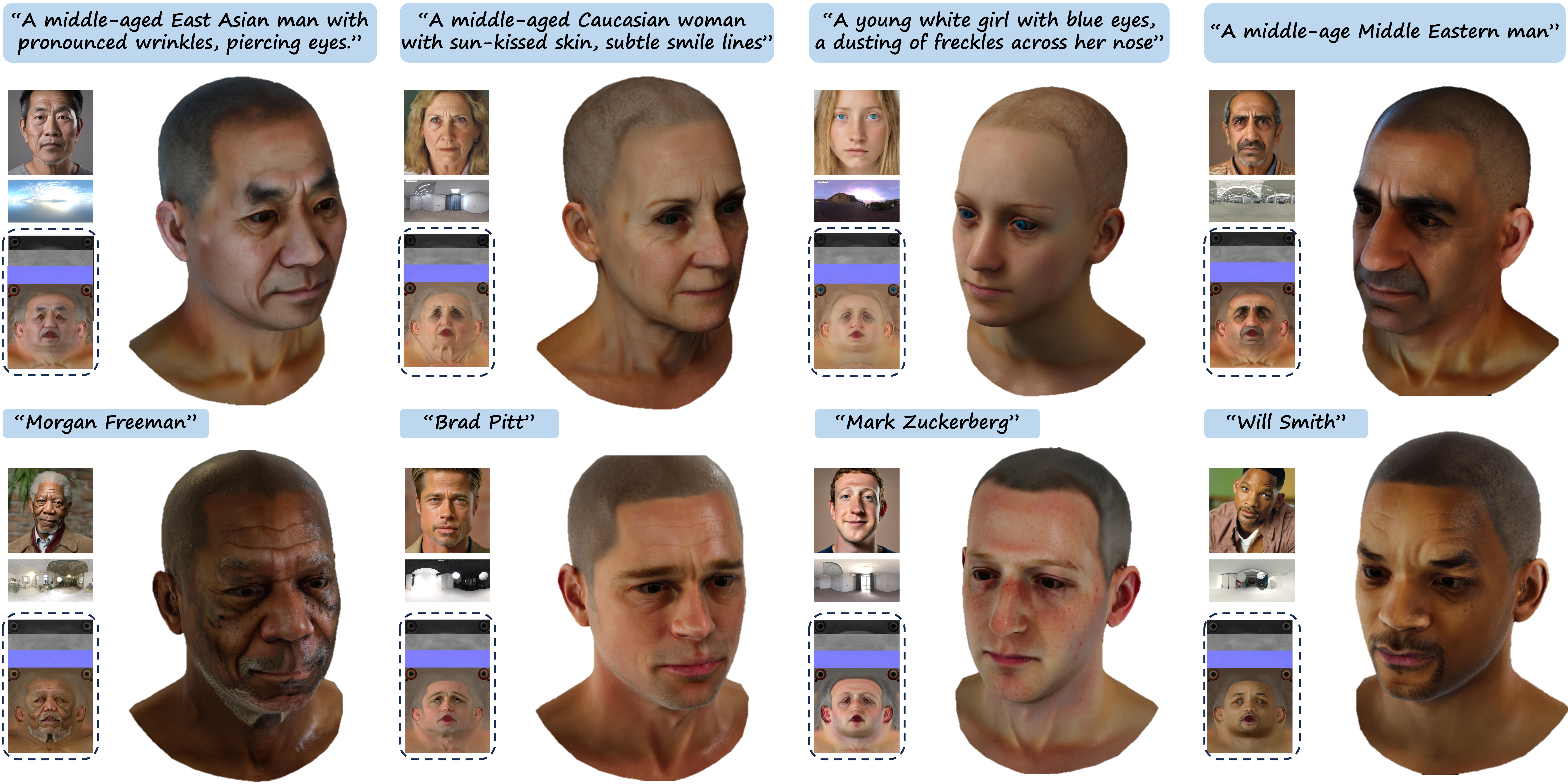}

   \caption{ \textbf{Results of generating random identities and celebrities.} We input the text prompts into the generic SDXL to create 2D face images. Our results showcase the reconstructed high-quality PBR textures which are also well-aligned with the meshes, exhibit high fidelity, and maintain the identity and facial details. To illustrate the quality of our generation, we relight each 3D avatar under various environment maps.  }
   \label{fig:results_1}
   \vspace{-3.00mm} 
\end{figure*}

\begin{figure}[t]
  \centering
   \includegraphics[width=0.94\linewidth]{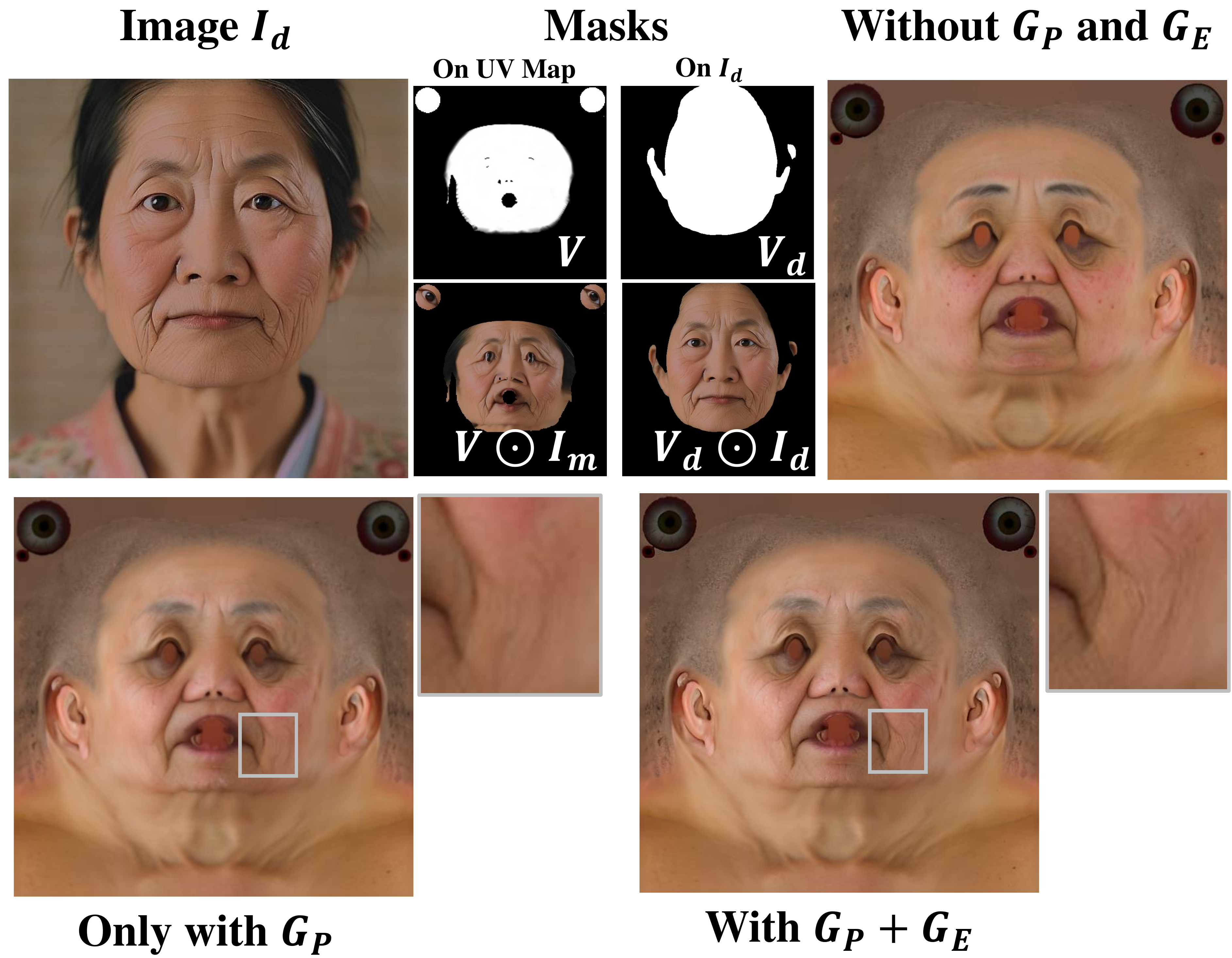}
   \caption{\textbf{Analysis of the guidances in the AGT-DM.} Three PBR textures generation scenarios from image $I_d$ by our AGT-DM are shown: one without $G_P$ and $G_E$, one only with $G_P$, and another with both $G_P$ and $G_E$. It clearly demonstrates that the identity and facial details are effectively maintained through these two guidances.}

   \label{fig:AGT}
   \vspace{-3.00mm} 
\end{figure}

\paragraph{Experimental Setup.}

We used SDXL\cite{podell2023sdxl} as our text-to-image generation model. We used pretrained BiSeNet \cite{yu2018bisenet,faceparsing} for generating face parsing mask, $S$. In our DCE module, we use the standard SD-2.1 base model and apply the DDIM sampling with 20 time steps. We preserve the res-features from $4^{th}$ to $11^{th}$ upsampling layers in the U-net extracted from the $I$, and inject them into the DDIM sampling of $z^I_T$, the query and key features from the $4^{th}$ to $9^{th}$ upsampling layers are extracted from DDIM inversion of $S$. We choose not to inject query and key features from all layers because we find injecting them to the last few layers can sometimes slightly change the identity. 

For our AGT-DM, we finetune the U-Net from SD-2.1-base model on our annotated 3DScan store dataset to generate diffuse color texture map. We attach ``A UV map of" to the text prompt during finetuning to generate FLAME UV maps. We train three decoders to output normal, specular and roughness maps. More training details are presented in the supplementary material.

In the AGT-DM, we use $T=200$, $N=90$, $\omega=7.5$, $\omega_p=0.1$, $\omega_{photo} = 0.4$, $\omega_{lpips} = 0.6$ and $\omega_e=0.05$. In the initial $T-N$ denoising steps, our approach adopts a latent space inpainting technique akin to the method described in \cite{avrahami2023blended}, utilizing a visibility mask. 
During the final $N$ steps, we apply the proposed photometric and edge guidances to rectify any misalignments and ensure a coherent integration between the observed and unobserved regions of the face. After the inference, we pass the resultant latent code to normal, specular and roughness decoders to obtain the corresponding PBR texture maps. We then pass the texture to a pre-trained Stable Diffusion super-resoltuion network \cite{rombach2022high} to get 2K resolution texture. 

\vspace{-3.00mm} 
\paragraph{Baselines.}

We show comparisons against different state-of-the-art approaches for text-to-avatar generation (Latent3d \cite{canfes2023text}, CLIPMatrix \cite{jetchev2021clipmatrix}, Text2Mesh\cite{michel2022text2mesh}, CLIPFace\cite{aneja2023clipface}, DreamFace\cite{zhang2023dreamface}) and image-to-avatar generation (FlameTex\cite{flametex}, PanoHead \cite{an2023panohead}) in the Table.~\ref{tab:method_comparison}. Details about the comparisons are included in the supplementary material.

\subsection{Results and Discussion}

We demonstrate our text/image generated realistic avatars in Fig.~\ref{fig:teaser} and ~\ref{fig:results_1}. Note that, we do not have those images in the training data for our AGT-DM. Generated results demonstrate rich textures maintaining fidelity to the given text prompt/image. Furthermore, due to our DCE model and AGT-DM's capabilities to extract diffuse color texture and PBR details, we can correctly render relighted avatars from any lighting condition. Since, AGT-DM enforces consistency across the observed and unobserved region, our rendered avatars look equally realistic from different angles without any visible artifacts.

\vspace{-3.00mm} 
\paragraph{Performance Analysis.}

For comparison, we randomly select 40 text prompts shown in the supplementary material, ensuring a comprehensive representation across various age groups, ethnicities and genders, as well as including a range of celebrities. For DreamFace and UltrAvatar, we render the generated meshes from 50 different angles under five different lighting conditions. For PanoHead, we provide five face images generated by SDXL for each text prompt, resulting in a total of 200 face images. Producing 50 different views for each prompt via PanoHead yields a total of 10k images (the same number applied to DreamFace and UltrAvatar). UltrAvatar can generate high-quality facial asset from text prompt within 2 minutes (compared to 5 minutes for DreamFace) on a single Nvidia A6000 GPU.

We evaluate the perceptual quality of the rendered images by using standard generative model metrics FID and KID. Similar to CLIPFace, we evaluate both of these metrics with respect to masked FFHQ images \cite{karras2019style}
(without background, eyes and mouth interior) as ground truth. For text-to-avatar generation, we additionally calculate CLIP score to measure similarity between text prompts and rendered images. We report the average score from two different CLIP variants, `ViT-B/16' and `ViT-L/14'.

Among the text-to-avatar generation approaches in Table~\ref{tab:method_comparison}, DreamFace performs very well on maintaining similarity between text and generated avatars. However, the generated avatars by DreamFace lack realism and diversity. Our proposed UltrAvatar performs significantly better than DreamFace in terms of perceptual quality (more results are shown in the supplementary material). Furthermore, in Fig.~\ref{fig:comparison}, we demonstrate that DreamFace fails to generate avatars from challenging prompts (e.g. big nose, uncommon celebrities). It is important to note that the results from DreamFace represent its best outputs from multiple runs.
 Our UltrAvatar also outperforms other text-to-avatar approaches in terms of perceptual quality and CLIP score, as reported in Table~\ref{tab:method_comparison}.

In the task of image-to-avatar generation, PanoHead achieves impressive performance in rendering front views. However, the effectiveness of PanoHead is heavily dependent on the accuracy of their pre-processing steps, which occasionally fail to provide precise estimation. Furthermore, the NeRF-based PanoHead approach is limited in relighting. Considering the multi-view rendering capabilities, UltrAvatar outperforms PanoHead in image-to-avatar task as shown in Table~\ref{tab:method_comparison}.

\begin{figure}[t]
   \centering 
    \includegraphics[width=0.7\linewidth]{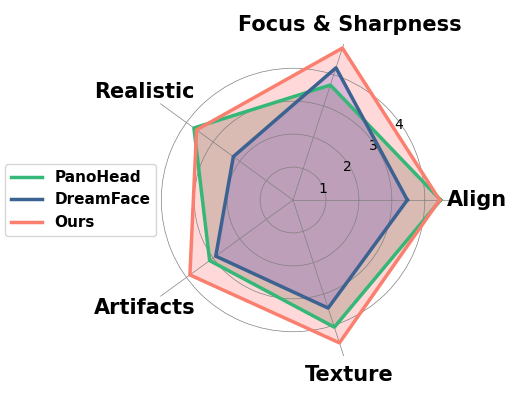}
    \caption{\textbf{Qualitative evaluation by GPT-4V.} Our framework has overall better performance.}
    \label{fig:gpt4}
    \vspace{-2.00mm} 
 \end{figure}

In addition, we automate text-to-avatar performance assessment utilizing GPT-4V(ision)~\cite{gpt4vblog, gpt4vcard}. GPT-4V is recognized for its human-like evaluation capabilities in vision-language tasks~\cite{zhang2023gpt4vision, yang2023dawn}. We evaluate models on a five-point Likert scale. The criteria for assessment include photo-realism, artifact minimization, skin texture quality, textual prompt alignment, and the overall focus and sharpness of the image. As illustrated in Fig.~\ref{fig:gpt4}, UltrAvatar demonstrates superior capabilities in generating lifelike human faces. It not only significantly reduces artifacts and enhances sharpness and focus compared to DreamFace and PanoHead but also maintains a high level of photo-realism and fidelity in text-prompt alignment.


\begin{table}[t]
\small
  \centering
  \begin{tabular}{|l|c|c|c|}
    \hline
    Method & FID $\downarrow$ & KID $\downarrow$ & CLIP Score $\uparrow$ \\
    \hline
    DreamFace~\cite{zhang2023dreamface} &  76.70 &  0.061 &  0.291 $\pm$ 0.020 \\
    ClipFace$^*$~\cite{aneja2023clipface} & 80.34 & 0.032 & 0.251 $\pm$ 0.059 \\
    Latent3d$^*$\cite{canfes2023text} & 205.27 & 0.260 & 0.227 $\pm$ 0.041 \\
    ClipMatrix$^*$\cite{jetchev2021clipmatrix}  & 198.34 & 0.180 & 0.243 $\pm$ 0.049 \\
    Text2Mesh$^*$\cite{michel2022text2mesh} & 219.59 & 0.185 & 0.264 $\pm$ 0.044 \\
    \hline
    FlameTex$^*$\cite{flametex} & 88.95 & 0.053 & - \\
    PanoHead~\cite{an2023panohead} &  48.64 & 0.039  &    -\\
    \hline
    UtrAvatar (Ours)   &  \textbf{45.50} & \textbf{0.029}  &  \textbf{0.301 $\pm$ 0.023}  \\
    \hline
  \end{tabular}
  \caption{Comparison of methods based on FID, KID, and CLIP Score metrics, \footnotesize{ $^*$ results are from CLIPFace.}}
  \label{tab:method_comparison}
  \vspace{-2.00mm}
\end{table}

\begin{figure}[t]
  \centering
   \includegraphics[width=0.9\linewidth]{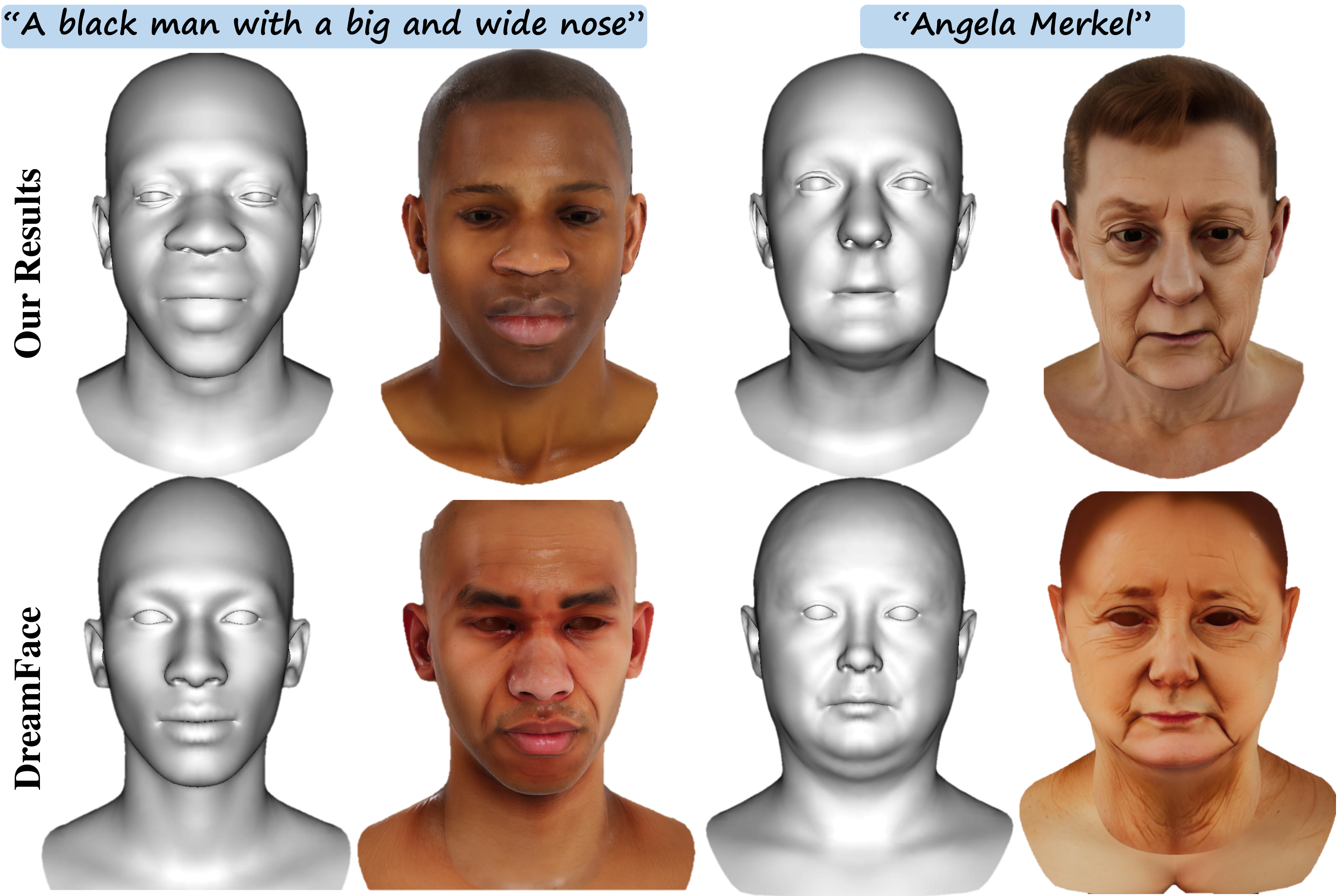}
   \caption{\textbf{Comparison to DreamFace.} Our results achieve better alignment with the text prompt than DreamFace, especially for extreme prompts.}
   \label{fig:comparison}
   \vspace{-4.00mm} 
\end{figure}



\subsection{Ablation Studies.}
\begin{figure}[t]
  \centering
   \includegraphics[width=1\linewidth]{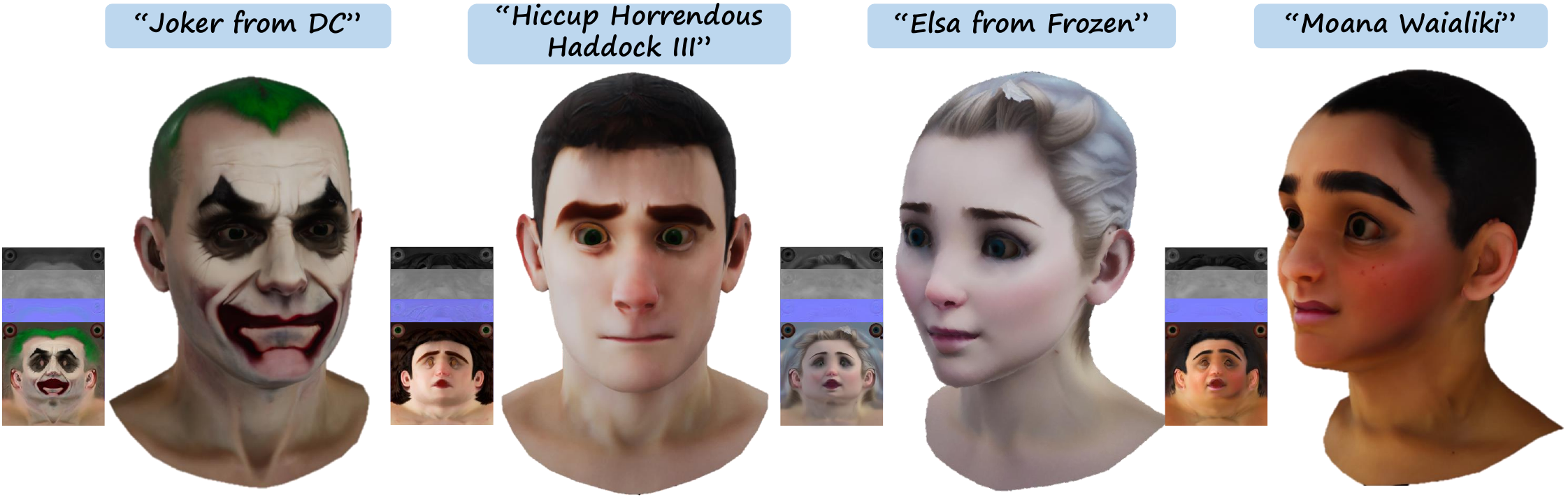}
   \caption{\textbf{Results of out-of-domain avatar generation.} Our framework has capability to generate out-of-distribution animation characters or non-human avatars.}
   \label{fig:nonhuman}
   \vspace{-4.00mm} 
\end{figure}

In Fig.~\ref{fig:AGT}, we illustrate the impact of different guidances on the AGT-DM performance. 
The photometric guidance enforces the similarity between the generated texture and the source image. Additionally, the edge guidance enhances the details in the generated color texture. 
\paragraph{Out-of-domain Generation.} UltrAvatar can generate avatars from the image/prompt of animation characters, comic characters and other out-of-domain characters. We have shown some results in Fig.~\ref{fig:nonhuman}.

\paragraph{Animation and Editing} Since our generated avatars are FLAME-based models, we can animate our generated avatars by changing the expressions and poses. We can also perform some texture editing using the text prompt in our AGT-DM. We have included the animation and editing results in the supplementary material.

\section{Conclusion}

We introduce a novel approach to 3D avatar generation from either a text prompt or a single image. At the core of our method is the DCE Model designed to eliminate unwanted lighting effects from a source image, as well as a texture generation model guided by photometric and edge signals to retain the avatar's PBR details. Compared with the other SOTA approaches, we demonstrate that our method can generate 3D avatars that display heightened realism, higher quality, superior fidelity and more extensive diversity.

{
    \small

}

\end{document}